\documentclass[preprint,12pt]{elsarticle}

\usepackage{amsmath,amsfonts,amssymb}
\usepackage{algorithmic}
\usepackage{algorithm}
\usepackage{array}
\usepackage[caption=false,labelfont=sf,textfont=sf]{subfig}
\usepackage{multirow}
\usepackage{textcomp}
\usepackage{stfloats}
\usepackage{url}
\usepackage{verbatim}
\usepackage{graphicx}
\usepackage{nicematrix,bm}
\usepackage{amsbsy,fixmath}
\usepackage{enumitem}
\usepackage{tabularx,ragged2e,multicol}
\usepackage[font=small,labelfont=bf]{caption}
\usepackage{subcaption,braket,mathtools}
\usepackage{rotating}

\journal{Journal}

\begin{document}

\begin{frontmatter}



\title{Efficient vectorized backpropagation algorithms for training feedforward networks composed of quadratic neurons}

\author[label1]{Mathew Mithra Noel\corref{cor1}}
\author[label1]{Venkataraman Muthiah-Nakarajan}
\author[label2]{Yug D Oswal}
\affiliation[label1]{organization={Vellore Institute of Technology},
             addressline={School of Electrical Engineering},
	         city={Vellore},
             postcode={632014},
             state={Tamil Nadu},
             country={India}}

\affiliation[label2]{organization={Vellore Institute of Technology},
             	addressline={School of Computer Science and Engineering},
             	city={Vellore},
             	postcode={632014},
             	state={Tamil Nadu},
             	country={India}}
\cortext[cor1]{Corresponding Author. Emails: mathew.m@vit.ac.in; mathew.mithra@gmail.com}

\begin{abstract}
Higher order artificial neurons whose outputs are computed by applying an activation function to a higher order multinomial function of the inputs have been considered in the past, but did not gain acceptance due to the extra parameters and computational cost. However, higher order neurons have significantly greater learning capabilities since the decision boundaries of higher order neurons can be quadric surfaces instead of just hyperplanes. This allows individual quadratic neurons to learn many nonlinearly separable datasets. Since quadratic forms can be represented by symmetric matrices, only $\frac{n(n+1)}{2}$ additional parameters are needed instead of $n^2$. A quadratic logistic regression model is first presented. Solutions to the XOR problem with a single quadratic neuron are considered. The complete vectorized equations for both forward and backward propagation in feedforward networks composed of quadratic neurons are derived. A reduced parameter quadratic neural network model with just $n$ additional parameters per neuron that provides a compromise between learning ability and computational cost is presented. Comparisons of benchmark classification datasets are used to demonstrate that a final layer of quadratic neurons enables networks to achieve higher accuracy with significantly fewer hidden layer neurons. In particular, this paper shows that any dataset composed of $\mathcal{C}$ bounded clusters can be separated with only a single layer of $\mathcal{C}$ quadratic neurons.				
\end{abstract}




\end{frontmatter}



\section{Introduction} 
The most common model of an artificial neuron is one in which the output of the neuron is computed by applying an affine function to the input. In particular the output or activation $a$ is computed using $ a = g(\mathbf{w}^T\mathbf{x}+b)$, where $g$ is the nonlinear activation function. The decision boundary of such a neuron is the set: 
$ B = \{\mathbf{x} \in \mathbb{R}^n:  \mathbf{w}^T\mathbf{x}+b = 0 \}$. This set $B$ is a hyperplane and hence can only separate linearly separable datasets. Thus each neuron in a traditional Artificial Neural Network (ANN) can only perform linear classification. In particular, single neurons cannot learn the XOR Boolean function. However, special pyramidal neurons capable of learning the XOR function have been recently discovered in the human neocortex which is responsible for higher order thinking \cite{Gidon}, \cite{m1}, \cite{m2}. This motivates the exploration of more complex artificial neuron models that can also individually learn the XOR function like biological neurons and potentially improve overall performance at the cost of a limited increase in complexity. 

The natural extension of the standard neuron model with hyperplane decision boundaries is to consider neurons that have quadric surfaces as decision boundaries. The output of a $2^{nd}$ order or quadratic neuron $a$ is computed using $a = g(\mathbf{x}^T\mathbf{Q}\mathbf{x}+\mathbf{w}^T\mathbf{x}+b)$, where $\mathbf{Q}$ is a symmetric matrix. Since $\mathbf{Q}$ is a symmetric matrix, only  $\frac{n(n+1)}{2}$ additional parameters are needed instead of $n^2$. In the past such higher order neurons have been considered but did not gain popularity due to the need for significantly more parameters, computational cost, lack of specialized ANN training hardware and efficient vectorized training algorithms. Although Backpropagation has been used to train Quadratic Neural Networks (QNNs) in the past, efficient vectorized forward and backpropagation equations have not been presented till now. In this paper, we derive the complete vectorized equations for forward and backpropagation in QNNs and show that QNNs can be trained efficiently. In particular it is shown that the computationally costly part of the calculations can be cached during forward propagation and reused during backpropagation. A reduced parameter QNN model that used only  $n$ parameters instead of  $\frac{n(n+1)}{2}$ additional parameters per neuron is also presented and the backpropagation algorithm in vectorized form is derived for this new model and shown to be computationally efficient.

In summary, the major contributions in this paper are:
\begin{enumerate}
	\item Elegant vectorized equations are derived for general QNNs (Section III)
	
	\item Elegant vectorized equations are derived for a new reduced parameter QNN (Section IV)
	
	\item A new quadratic logistic-regression model that allows single neuron solutions to the famous XOR problem (Section II)
	
	\item Single layer QNNs are proven to solve problems impossible with single layer ANNs of arbitrary size (Results: Section A)
	
	\item Proof that computationally expensive calculations in Forward Propagation can be reused during backpropagation in QNNs and RPQNNs (Algorithm 1 \& 2)
	
	\item A comparison of the computational complexity of ANN and QNNs (Section V)
	
	\item Asymptotic computational complexity of ANN and RPQNN are shown to be the same namely $\mathcal{O}(n^2)$ (Section V)
	
	\item Asymptotic  computational complexity of QNN is shown to be $\mathcal{O}(n^3)$, same as Gaussian Elimination (Section V)
	
\end{enumerate}

QNNs are yet to gain widespread acceptance, so the literature on QNNs is limited. In the following we present a survey of major contributions to QNN research.

\subsection*{Literature survey}
Higher order neural (HON) networks were investigated for their increased flexibility since the 1970s \cite{ivanko} \cite{strokes} \cite{giles}, but failed to gain popularity due to the unavailability of high performance computing hardware, large datasets and efficient algorithms. Giles et al. explored learning behaviour and  overfitting in HONs \cite{giles}. The greater suitability of QNNs compared to standard ANNs for hardware VLSI implementations was described in \cite{hardware}. Alternatives to the BP algorithm for training QNNs was investigated in \cite{AQNN}. Despite the lack of popularity of QNNs due to their perceived computational complexity, many successful applications of QNNs have been reported. \cite{gaussian} reports on the superiority of QNNs over conventional ANNs for classification of gaussian mixture data in the recent past. An exploration of the possible advantages of QNNs is presented in \cite{QNN}. An improvement in the accuracy of Convolutional Neural Networks (CNNs) with quadratic neurons on image classification tasks was reported in \cite{CQNN} \cite{QNN2}. The unique features of HON networks are described in \cite{HON2}. Applications of higher order recurrent neural networks for nonlinear control and system identification are explored in \cite{alanis}, \cite{russ} and \cite{ivo}.

CNNs and QNNs serve completely different roles and hence are not comparable. CNNs are inspired my mammalian visual cortex and serve as very effective feature extractors for image data. The features extracted by the initial convolutional layers are then processed by FNN (fully connected) layers in a typical CNN model. QNNs are generalizations of FNNs and hence are directly comparable to FNNs. Given that convolutional layers are already very effective as feature extractors, the extra parameters introduced by quadratic neurons are not justified and hence the authors do not propose replacing convolutional layers with QNN layers in convolutional layers. Rather, the final fully connected and Softmax layers in a standard ANN or CNN can be replaced with QNN or RPQNN to achieve performance benefits without adding many extra parameters. 

We begin our exploration of QNNs by considering logistic regression with a single quadratic neuron in some detail next to understand possible advantages and limitations in a simple setting.

\section{Quadratic Logistic Regression}
In the following, vectorized Stochastic Gradient Descent (SGD) update equations for logistic regression with a single quadratic neuron are presented. In the standard logistic regression model that uses a single sigmoidal neuron, the goal is to learn a hyperplane that separates the classes. The quadratic logistic regression model proposed in this paper generalizes the standard logistic regression model by learning a hyper-quadric surface ($\mathbf{x}^T\mathbf{Q}\mathbf{x}+\mathbf{w}^T\mathbf{x}+b = 0$) that separates the dataset. The variables associated with a quadratic logistic regression model are:

\begin{eqnarray}
	\text{Target or class label    }	y&\in&\{0,1\} \nonumber \\
	\text{Input vector    }\mathbf{x}&\in&\mathbb{R}^d \nonumber \\
	\text{Output    }	\hat{y} &\in&(0,1)  \nonumber\\
	\text{Weight vector    }	\textbf{w} &\in&\mathbb{R}^d \nonumber \\
	\text{Bias parameter    }	b &\in& \mathbb{R} \nonumber \\
	\text{Symmetric parameter matrix   }\mathbf{Q} &\in& \mathbb{R}^{d\times d} \nonumber
\end{eqnarray}

The output is calculated by applying the logistic sigmoid activation function to a general quadratic function of the inputs as follows:

\begin{align}\label{model}
	\begin{split}
		\hat{y}					&= \sigma(\mathbf{x}^T\mathbf{Q}\mathbf{x}+\mathbf{w}^T\mathbf{x}+b) \\
		&= \sigma(z)
	\end{split}  
\end{align}

Where $z=\mathbf{x}^T\mathbf{Q}\mathbf{x}+\mathbf{w}^T\mathbf{x}+b$ and 

\begin{equation} \nonumber
	\sigma(z)=\frac{1}{1+e^{-z}}  
\end{equation}

It is well known that the derivative of the sigmoid can be expressed in terms of its output.

\begin{eqnarray} 
	\sigma^\prime(z)&=&\sigma(z)(1-\sigma(z))
\end{eqnarray}

The loss function for the binary classification task is:

\begin{eqnarray}
	l(y,\hat{y})=-[y\ln{\hat{y}}+(1-y)\ln{(1-\hat{y})}] 
\end{eqnarray}

To perform parameter updates using SGD, the following partial derivates are needed: $\frac{\partial l}{\partial b}$, $\frac{\partial l}{\partial w_i}$, $\frac{\partial l}{\partial q_{ij}}$. These partial derivatives can be computed using the "Chain Rule" from calculus (\ref{chain}).

\begin{align} \label{chain}
	\begin{split}
		\frac{\partial l}{\partial b} &= \frac{\partial l}{\partial \hat{y}}.\frac{\partial \hat{y}}{\partial z}.\frac{\partial z}{\partial b}  \\
		\frac{\partial l}{\partial w_i} &= \frac{\partial l}{\partial \hat{y}}.\frac{\partial \hat{y}}{\partial z}.\frac{\partial z}{\partial w_i}  \\
		\frac{\partial l}{\partial q_{ij}}& =\frac{\partial l}{\partial \hat{y}}.\frac{\partial \hat{y}}{\partial z}.\frac{\partial z}{\partial q_{ij}} 
	\end{split}
\end{align}

The derivatives needed to compute $\frac{\partial l}{\partial b}$ in (\ref{chain}) are computed in (\ref{chain1}):

\begin{align} \label{chain1}
	\begin{split}
		\frac{\partial l}{\partial \hat{y}}&=\frac{\hat{y}-y}{\hat{y}(1-\hat{y})}  \\
		\frac{\partial \hat{y}}{\partial z}&= \hat{y}(1-\hat{y}) \\
		\frac{\partial z}{\partial b}&=1 
	\end{split}
\end{align}

Finally, $\frac{\partial l}{\partial b}$ can be obtained from (\ref{chain}) and (\ref{chain1}):

\begin{align} \label{chain2}
	\begin{split}
		\frac{\partial l}{\partial \hat{y}}\cdot\frac{\partial \hat{y}}{\partial z}&=\hat{y}-y  \\
		\implies	\frac{\partial l}{\partial b}&=(\hat{y}-y) 	
	\end{split} 
\end{align}

The standard SGD parameter update rule for any parameter $\Theta$ is:

\begin{equation} \label{sgd1}
	\Theta\leftarrow\Theta - \eta\frac{\partial l}{\partial\Theta}
\end{equation}

From (\ref{chain2}) and (\ref{sgd1}), the SGD update rule for parameter $b$ is:

\begin{equation} \label{bupdate}
	b\leftarrow b-\eta \frac{\partial l}{\partial b}=b+\eta(y-\hat{y})
\end{equation}

From (\ref{model}) we note that: 

\begin{equation} \label{w}
	\frac{\partial z}{\partial w_i} = x_i
\end{equation}		

The SGD update for $w_i$ is now obtained from (\ref{chain}), (\ref{chain2}), (\ref{sgd1}) and (\ref{w}):

\begin{eqnarray} 
	w_i&\leftarrow& w_i-\eta \frac{\partial l}{\partial w_i} \nonumber\\
	w_i&\leftarrow&w_i+\eta (y-\hat{y})x_i \label{wupdate}
\end{eqnarray}

The above weight update equations can be expressed in vectorized notation (\ref{w_update}).

\begin{equation}
	\textbf{w}\leftarrow \textbf{w} - \eta(y-\hat{y})\textbf{x} 
	\label{w_update}
\end{equation}

Next we derive the SGD update rule for $q_{ij}$. From (\ref{model}) we note that:

\begin{equation}  \nonumber
	\frac{\partial z}{\partial q_{ij}}=\frac{\partial}{\partial q_{ij}}(\textbf{x}^T\mathbf{Q}\mathbf{x}+\mathbf{w}^T\mathbf{x}+b)  
\end{equation}

Since $\mathbf{w}^T\mathbf{x}+b$ does not depend on $q_{ij}$:

\begin{eqnarray}
	\frac{\partial z}{\partial q_{ij}}	&=&\frac{\partial}{\partial q_{ij}}(\mathbf{x}^T\mathbf{Q}\mathbf{x})  \nonumber\\
	&=& \frac{\partial}{\partial q_{ij}}\left(\sum_{l=1}^{d}\sum_{m=1}^{d}q_{lm}x_lx_m\right) \label{qresult}
\end{eqnarray}

(\ref{qresult}) can be simplified to obtain (\ref{ela}) below: 

\begin{equation}
	\frac{\partial z}{\partial q_{ij}}=\left\{ \begin{tabular}{lcr}
		$\frac{\partial}{\partial q_{ij}}(q_{ij}x_ix_j+q_{ji}x_jx_i)$ & if& $i\neq j$ \\
		$\frac{\partial}{\partial q_{ii}}(q_{ii}x_i^2)$ & if&  $i=j$ 
	\end{tabular} \right. \label{ela}
\end{equation}

(\ref{ela}) can be further simplified to yield (\ref{red}) below.

\begin{equation} \label{red}
	\frac{\partial z}{\partial q_{ij}}=\left\{\begin{tabular}{lr}
		$2x_ix_j$& if $i \neq j$ \\
		$x_i^2$ & if $i=j$
	\end{tabular}\right.
\end{equation}

These partial derivatives can be collected together for notational convenience and computational efficiency using the concept of a derivative with respect to a matrix. Given a function $ f:\mathbb{R}^{m \times n}\rightarrow \mathbb{R} $,  $ \frac{\partial f}{\partial A} $ is defined using $ \frac{\partial f}{\partial A} \coloneqq    \left[\frac{\partial f}{\partial A_{ij}}\right]    $. Thus, $\frac{\partial (\mathbf{x}^T\mathbf{Q}\mathbf{x})}{\partial \mathbf{Q}}$ is the matrix given in (\ref{bmatrix}) below.

\begin{eqnarray}
	\frac{\partial (\mathbf{x}^T\mathbf{Q}\mathbf{x})}{\partial \mathbf{Q}}&\triangleq&\begin{bmatrix}
		x_1^2 & 2x_1x_2  & \cdots & 2x_1x_d \\
		2x_2x_1& x_2^2  & \cdots &  2x_2x_d \\
		\vdots & \vdots &  \ddots & \vdots \\
		2x_dx_1 & 2x_dx_2 & \cdots & x_d^2
	\end{bmatrix} \label{bmatrix} \\
	&\triangleq& \mathbf{M}(\mathbf{x})  
\end{eqnarray}

The SGD update for $q_{ij}$ is:

\begin{equation}
	q_{ij}\leftarrow q_{ij}-\eta \frac{\partial l}{\partial q_{ij}} \label{qupdate} 
\end{equation}

From (\ref{chain2}), (\ref{red}) and (\ref{qupdate})
\begin{eqnarray}
	q_{ij}	&\leftarrow&q_{ij}-\eta (\hat{y}-y)\frac{\partial z}{\partial q_{ij}} \nonumber \\
	q_{ij}	&\leftarrow& \left\{ \begin{tabular}{lcr}
		$q_{ij}+\eta(y-\hat{y})2x_ix_j$ &if& $i\neq j$ \\
		$q_{ij}+\eta(y-\hat{y})x_i^2$&if& $i=j$
	\end{tabular}\right. \label{qup}
\end{eqnarray}

The above update equations (\ref{qup}) for $q_{ij}$ can be compactly expressed in vectorized form using (\ref{bmatrix}) to obtain (\ref{Q_update}).

\begin{equation} 			\label{Q_update}
	\mathbf{Q}\leftarrow \mathbf{Q}+\eta(y-\hat{y})\mathbf{M}(\mathbf{x})
\end{equation}

Equations (\ref{bupdate}), (\ref{w_update}) and (\ref{Q_update}) are the vectorized parameter update equations for training a quadratic logistic regression model. Next we show that a single quadratic neuron can learn the XOR function.

\subsection*{Single neuron solutions to the XOR problem}

The XOR problem is the task of learning the XOR dataset shown below in (\ref{Eq: XOR dataset}). For mathematical convenience the boolean variables 0 and 1 are encoded as $-1$ and $1$, respectively (bipolar encoding).

\begin{equation}\label{Eq: XOR dataset}
	D = \left\{ 
	\left(\begin{bmatrix} -1\\-1 \end{bmatrix}, -1\right), \left(\begin{bmatrix} 1\\-1 \end{bmatrix}, 1\right),
	\left(\begin{bmatrix} -1\\1 \end{bmatrix}, 1\right),
	\left(\begin{bmatrix}  1\\1 \end{bmatrix}, -1\right)
	\right\}
\end{equation}

\begin{figure*}
	\centering
	\subfloat[]{\includegraphics[width=0.54\textwidth]{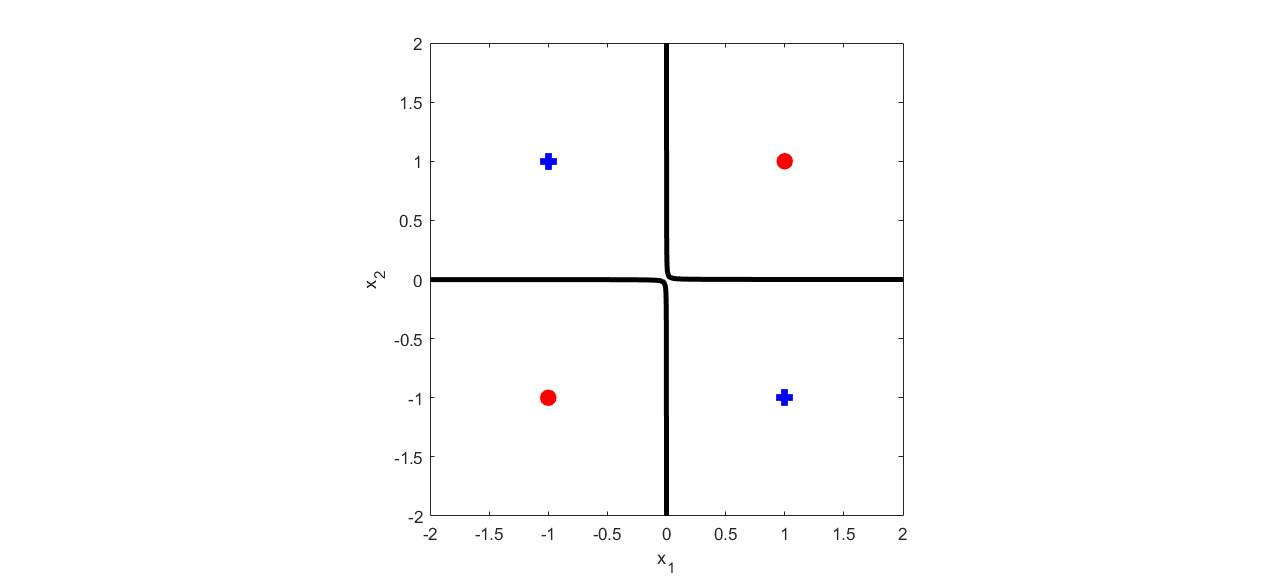}}
	\subfloat[]{\includegraphics[width=0.54\textwidth]{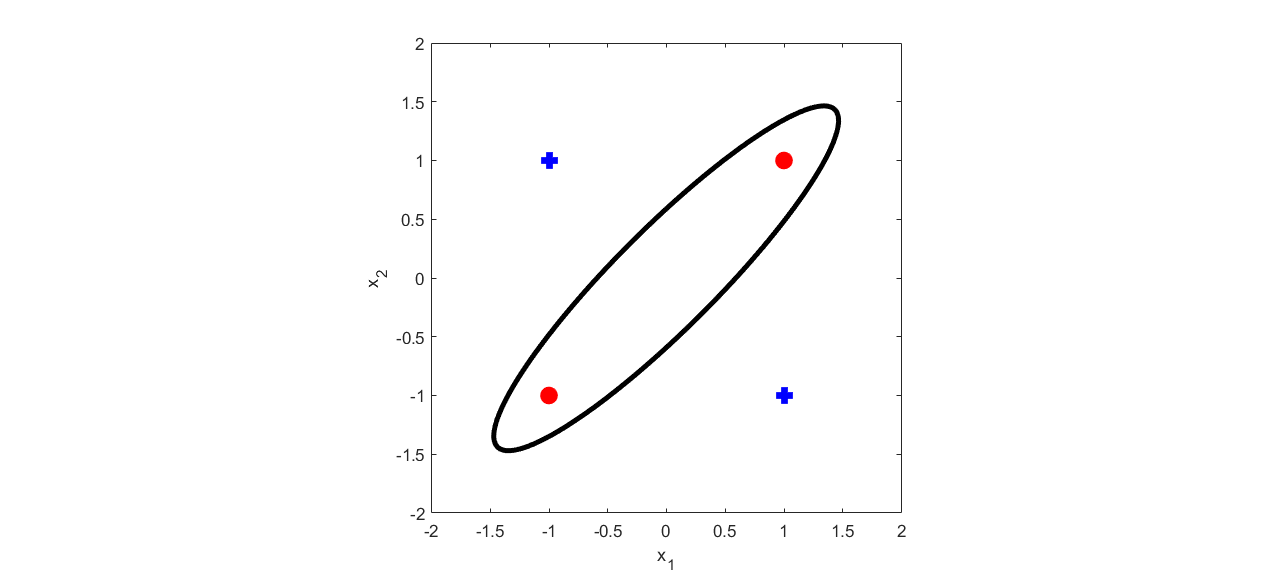}}
	\caption{A single quadratic neuron is able to separate the XOR dataset with a hyperbola or an ellipse. Two possible solutions are shown. (a) When $\mathbf{Q}$ is initalized with a random matrix, a hyperbolic decision boundary is obtained. (b) When $\mathbf{Q}$ is initalized with the Identity matrix, an ellipsoidal decision boundary is obtained.}
	\label{xor}			
\end{figure*}

Using (\ref{bupdate}), (\ref{w_update}) and (\ref{Q_update}), a single quadratic neuron can be trained to learn the XOR function. From Fig. \ref{xor} it is clear that a single quadratic neuron can learn complex quadric surfaces (conic sections in 2D) to separate nonlinearly separable datasets. Fig. \ref{xor} (a) shows one solution to the XOR problem where the XOR dataset is separated by a hyperbolic decision boundary. Fig. \ref{xor} (b) shows another solution to the XOR problem where the XOR dataset is separated by an elliptic decision boundary. The elliptic decision boundary was obtained by initializing $\mathbf{Q}$ with the positive definite identity matrix and the hyperbolic decision boundary was obtained when the $\mathbf{Q}$ matrix was initialized with a random matrix.

In the following we consider general feedforward networks consisting of multiple layers of quadratic neurons and derive the vectorized BP algorithm update equations.

\section{Backpropagation in feedforward artificial neural networks with quadratic neurons}

The following notation is used to represent a QNN model.

\begin{eqnarray}
	z_k^l &=& \mbox{Total input to the $k$-th neuron in the $l^{th}$ layer} \nonumber \\
	w^l_{ki} &=& \mbox{Weight parameter connecting the $i^{th}$ neuron in the} \nonumber \\
	& & \mbox{$(l-1)^{th}$ layer to the $k^{th}$ neuron in the $l^{th}$ layer } \nonumber \\
	a^l_k &=&\mbox{Output of the $k^{th}$ neuron in the $l^{th}$ layer}  \nonumber\\
	a_k^l &=& g_l(z^l_k); \mbox{ where $g_l$ is the activation function used} \nonumber \\ && \mbox{ in the $l^{th}$ layer} \nonumber
\end{eqnarray}

The quadratically weighted input to the $k$-th neuron in the $l^{th}$ layer is:

\begin{eqnarray}
	z^l_k&=&b^l_k+\sum_{i=1}^{n_{l-1}} w^l_{ki}a^{l-1}_i+\sum_{m=1}^{n_{l-1}}\sum_{n=1}^{n_{l-1}}q_{mn}^{lk}a^{l-1}_ma^{l-1}_n \label{fd} \\
	&=&b^l_k+W^l_{k:}\mathbf{a}^{l-1}+(\mathbf{a}^{l-1})^T Q^{lk}\mathbf{a}^{l-1} \label{vecfd}
\end{eqnarray}

Where $b^l_k$ are the bias parameters of the $k^{th}$ neuron in the $l^{th}$ layer, $W^l_{k:}$ is the $k^{th}$ row vector of $W^l$ ($l^{th}$ layer weight matrix) and $\mathbf{a}^{l-1}$ is the vector of outputs from the $(l-1)^{th}$ layer.

Equation (\ref{vecfd}) can be concisely expressed in vectorized form (\ref{eq3}).

\setlength{\arraycolsep}{0.05em}
\begin{eqnarray}
	\mathbf{z}^l&=&\mathbf{b}^l+\mathbf{W}^l\mathbf{a}^{l-1} \nonumber\\
	&+& \begin{bmatrix}
		(\mathbf{a^{l-1})}^T & 0 & \cdots & 0 \\
		0 & (\mathbf{a^{l-1})}^T & \cdots & 0 \\
		\vdots & \vdots & \ddots & \vdots \\
		0 & 0 & \cdots & (\mathbf{a^{l-1})}^T 
	\end{bmatrix} \begin{bmatrix}
		Q^{l1}\\
		Q^{l2}\\
		\vdots \\
		Q^{ln_l}
	\end{bmatrix}\mathbf{a}^{l-1}\nonumber\\
	&=& \mathbf{b}^l + (\mathbf{W}^l+\mathbf{A}^{l-1}\mathbf{Q}^l)\mathbf{a}^{l-1} \label{eq3}
\end{eqnarray}
\setlength{\arraycolsep}{5pt}

Where $Q^{lk}=[q_{mn}^{lk}]$ is the matrix of parameters associated with the quadratic term for the $k^{th}$ neuron in the $l^{th}$ layer. The individual $Q^{lk}$ matrices in the $l^{th}$ layer are collected in a single block matrix $\mathbf{Q}^l$ for convenience.

Based on the above, the equations for forward propagation in a QNN are summarized in (\ref{FP}).

\begin{align} \label{FP}
	\begin{split}
		\mathbf{z}^l&=\mathbf{b}^l+(\mathbf{W}^l+\mathbf{A}^{l-1}\mathbf{Q}^l)\mathbf{a}^{l-1} \\
		\mathbf{a}^l&=g_l(\mathbf{z}^l)\mbox{ where } l=1,2,3,\cdots,L
	\end{split}
\end{align}

For generality and simplicity, the non-mutually exclusive multi-label classification task is considered. Cross-entropy (\ref{loss}) is the standard loss function for multi-label classification tasks and we use the same.

\begin{equation} \label{loss}
	\mathcal{L}(\mathbf{y},\mathbf{\hat{y}})=-\sum_{p=1}^{n_L}\left[y_p\ln{\hat{y}_p}+(1-y_p)\ln{(1-\hat{y}_p)}\right]
\end{equation}

The free parameters in the above QNN model are $b^l_k$, $w_{kj}^l$ and $q_{mn}^{lk}$ and which is represented by $\Theta$.

Using the Chain Rule:

\begin{equation}\label{update}
	\frac{\partial \mathcal{L}}{\partial \Theta} =\frac{\partial \mathcal{L}}{\partial z^l_k}.\frac{\partial z^l_k}{\partial \Theta}=\delta^l_k\frac{\partial z^l_k}{\partial \Theta}.
\end{equation}

Since

\begin{equation}\label{Db}
	\frac{\partial z^l_k}{\partial b^l_k}=1 \implies \frac{\partial L}{\partial b^l_k}=\delta^l_k
\end{equation}

(\ref{Db}) is written in vectorized form below.

\begin{equation} \label{paraup1}
	\frac{\partial \mathcal{L}}{\partial \mathbf{b}^l} = \bm{\delta}^l
\end{equation}

Applying the chain rule again results in (\ref{one}) and (\ref{two}):

\begin{equation} \label{one}
	\frac{\partial \mathcal{L}}{\partial w^l_{kj}}=\delta^l_ka^{l-1}_j
\end{equation}

\begin{equation} \label{two}
	\frac{\partial z^l_k}{\partial w^l_{kj}}=a^{l-1}_j
\end{equation}

The above results can be expressed in vectorized form (\ref{thr}).

\begin{eqnarray}
	\frac{\partial \mathcal{L}}{\partial \mathbf{W}^l}&=& \left[\frac{\partial \mathcal{L}}{\partial w^l_{pq}}\right]=[\delta^l_p a^{l-1}_q] \nonumber\\
	&=& \bm{\delta}^l(\mathbf{a}^{l-1})^T= \bm{\delta}^l\otimes\mathbf{a}^{l-1} \label{thr}
\end{eqnarray}

The update for $\mathbf{Q}$ can be derived starting from (\ref{update}).

\begin{equation} \label{uti4}
	\frac{\partial \mathcal{L}}{\partial q^{lk}_{rs}}=\delta^l_k\frac{\partial z^l_k}{\partial q^{lk}_{rs}}
\end{equation}

where, 

\begin{eqnarray}
	\frac{\partial z^l_k}{\partial q_{rs}^{lk}}&=& \frac{\partial}{\partial q^{lk}_{rs}}\left[\sum_{m=1}^{n_l}\sum_{n=1}^{n_l}q^{lk}_{mn}a_m^{l-1}a_n^{l-1}\right] \nonumber \\ 
	&=& \left\{\begin{tabular}{lcr} \label{uti2}
		$2a^{l-1}_ra^{l-1}_s$ &if & $r\neq s$ \\
		$(a_r^{l-1})^2$&if& $r=s$\end{tabular}\right. \nonumber \\
	&\triangleq& [m^{lk}_{rs}] \label{uti1}
\end{eqnarray}

Substituting (\ref{uti1}) in (\ref{uti4})

\begin{equation} \label{utilize}
	\frac{\partial \mathcal{L}}{\partial Q^{lk}} =\delta^l_kM^{lk}
\end{equation}

\begin{eqnarray}
	M^{lk} &=&\begin{bmatrix}
		(a^{l-1}_1)^2 & 2a_1^{l-1}a_2^{l-1} & \cdots& 2a_1^{l-1}a_{n_l}^{l-1} \\
		2a^{l-1}_2a^{l-1}_1& (a_2^{l-1})^2 & \cdots& 2a^{l-1}_2a^{l-1}_{n_l} \\
		\vdots & \vdots & \ddots & \vdots \\
		2a^{l-1}_{n_l}a^{l-1}_1 & 2a^{l-1}_{n_l}a^{l-1}_2&\cdots & (a^{l-1}_{n_l})^2
	\end{bmatrix} 
\end{eqnarray}

The matrix $M^{lk}$ is independent of $k$ and depends only on the outputs from the previous layer $\mathbf{a}^{l-1}$. Thus $M^{lk}=\mathbf{M}^l(\mathbf{a}^{l-1})$ where $k=1,2,\cdots,n_l$. Equation (\ref{utilize}) for every $l$ and $k$ can be collected together and concisely written in computationally efficient vectorized form using the Kronecker product (\ref{fourtyone}). 

\begin{equation} \label{fourtyone}
	\begin{bmatrix}
		\frac{\partial \mathcal{L}^l}{\partial Q^{l1}} \\
		\frac{\partial \mathcal{L}}{\partial Q^{l2}} \\
		\vdots \\
		\frac{\partial \mathcal{L}}{\partial Q^{ln_l}}
	\end{bmatrix} = 
	\begin{bmatrix}
		\delta_1^l \mathbf{M}^l(\mathbf{a}^{l-1}) \\
		\delta_2^l \mathbf{M}^l(\mathbf{a}^{l-1}) \\
		\vdots  \\
		\delta_{n_l}^l \mathbf{M}^l(\mathbf{a}^{l-1}) 
	\end{bmatrix} = \bm{\delta}^l \circledast \mathbf{M}^l(\mathbf{a}^{l-1})
\end{equation}					

In (\ref{fourtyone}), $\circledast$ is the Kronecker product. Recognizing that the left hand side expression in (\ref{fourtyone}) is just $\frac{\partial \mathcal{L}}{\mathbf{Q}^l}$, we obtain an elegant expression for the derivatives of all the additional parameters associated with a QNN layer (\ref{fourtwo}).

\begin{equation} \label{fourtwo}
	\frac{\partial \mathcal{L}}{\partial \mathbf{Q}^l}= \bm{\delta}^l \circledast \mathbf{M}^l(\mathbf{a}^{l-1})
\end{equation} 

Now, we consider backpropagating the error to compute $\delta^{l-1}$ from  $\delta^{l}$.

\begin{eqnarray}
	\delta^{l-1}_t &=& \frac{\partial \mathcal{L}}{\partial z_t^{l-1}} \nonumber\\
	&=& \sum_{k=1}^{n_l} \frac{\partial \mathcal{L}}{\partial z^l_k}.\frac{\partial z^l_k}{\partial a^{l-1}_t}.\frac{\partial a^{l-1}_t}{\partial z^{l-1}_t} \nonumber\\
	&=&\sum_{k=1}^{n_l}\delta^l_k\frac{\partial z^l_k}{\partial a^{l-1}_t}g_{l-1}^\prime(z^{l-1}_t)  \label{eq6}
\end{eqnarray}

Since,

\begin{equation} \nonumber
	z^l_k=b^l_k+\sum_{i=1}^{n_l}w^l_{ki}a^{l-1}_i+\sum_{m=1}^{n_l}\sum_{n=1}^{n_l}q^{lk}_{mn}a^{l-1}_ma^{l-1}_n
\end{equation}

\begin{equation} \label{eq7}
	\frac{\partial z^l_k}{\partial a^{l-1}_t}=w^l_{kt}+2\sum_p^{n_l} q^{lk}_{tp}a^{l-1}_p
\end{equation}

Substituting (\ref{eq7}) in (\ref{eq6}), we get:

\begin{eqnarray}
	\delta^{l-1}_t&=&\sum_{k=1}^{n_l}\delta_k^lw^l_{kt}g_{l-1}^\prime(z^{l-1}_t)+2\sum_{p=1}^{n_l} (q^{lk}_{tp}a^{l-1}_p)g_{l-1}^\prime(z^{l-1}_t)\nonumber\\
	&=& g_{l-1}^\prime(z^{l-1}_t)\left[\sum_{k=1}^{n_l}(W^l_{tk})^T \delta_k^l +2\sum_{k=1}^{n_l}\delta_k^l\sum_{p=1}^{n_l} q^{lk}_{tp}a^{l-1}_p \right]  \label{delta}
\end{eqnarray}

(\ref{delta}) is presented in vectorized form in (\ref{deltav}).

\begin{equation}
	\bm{\delta}^{l-1} = g_{l-1}^\prime(\mathbf{z}^{l-1})\odot\left[ (\mathbf{W}^l)^T\bm{\delta}^l+2\sum_{k=1}^{n_l}\delta_k^lQ^{lk}\mathbf{a}^{l-1}\right]  \label{deltav}
\end{equation}

Now, the second term inside the square brackets of (\ref{deltav}) is,

\begin{equation}
	\sum_{k=1}^{n_l}\delta_k^l(Q^{lk}\mathbf{a}^{l-1})=\begin{bmatrix}
		Q^{l1}\mathbf{a}^{l-1} &Q^{l2}\mathbf{a}^{l-1}&\cdots &Q^{ln_l}\mathbf{a}^{l-1}
	\end{bmatrix} \bm{\delta}^l \label{eq9}
\end{equation}

Similar to (\ref{eq3}):
\begin{equation} \label{new1}
	\begin{bmatrix}
		(\mathbf{a}^{l-1})^TQ^{l1}\mathbf{a}^{l-1}\\
		(\mathbf{a}^{l-1})^TQ^{l2}\mathbf{a}^{l-1} \\
		\vdots \\
		(\mathbf{a}^{l-1})^TQ^{ln_l}\mathbf{a}^{l-1}
	\end{bmatrix}=\mathbf{A}^{l-1}\mathbf{Q}^{l}\mathbf{a}^{l-1} 
\end{equation}

From (\ref{new1}), we note the following:

\setlength{\arraycolsep}{0.00em}
\begin{eqnarray}
	\mathbf{A}^{l-1}\mathbf{Q}^l&=&\begin{bmatrix}
		(\mathbf{a}^{l-1})^T&0&\cdots&0 \\
		0& (\mathbf{a}^{l-1})^T& \cdots& 0 \\
		\vdots&\vdots&\ddots & \vdots\\
		0&0&\cdots&(\mathbf{a}^{l-1})^T
	\end{bmatrix}\begin{bmatrix}
		Q^{l1}\\
		Q^{l2}\\
		\vdots\\
		Q^{ln_l}
	\end{bmatrix}\nonumber\\
	&=&
	\begin{bmatrix}
		(\mathbf{a}^{l-1})^TQ^{l1} \\
		(\mathbf{a}^{l-1})^TQ^{l2} \\
		\vdots \\
		(\mathbf{a}^{l-1})^TQ^{ln_l}
	\end{bmatrix} \label{new2}
\end{eqnarray}
\setlength{\arraycolsep}{5pt}
The transpose of (\ref{new2}) is computed in (\ref{eq10})
\begin{eqnarray}
	(\mathbf{A}^{l-1}\mathbf{Q}^l)^T &=& (\mathbf{Q}^l)^T(\mathbf{A}^{l-1})^T\nonumber \\
	&=& \begin{bmatrix}
		Q^{l1}\\Q^{l2}\\ \vdots\\Q^{ln_l}
	\end{bmatrix}^T
	\begin{bmatrix}
		\mathbf{a}^{l-1}&0&\cdots&0\\
		0&\mathbf{a}^{l-1}&\cdots&0\\
		\vdots&\vdots&\ddots& 0\\
		0&0&0&\mathbf{a}^{l-1}
	\end{bmatrix}\nonumber
\end{eqnarray}
\begin{equation} 
	(\mathbf{A}^{l-1}\mathbf{Q}^l)^T=\begin{bmatrix}
		Q^{l1}\mathbf{a}^{l-1}&Q^{l2}\mathbf{a}^{l-1}&\cdots& Q^{ln_l}\mathbf{a}^{l-1}
	\end{bmatrix}	\label{eq10}
\end{equation}

From (\ref{deltav}), (\ref{eq9}) and (\ref{eq10}), we obtain the vectorized backpropagation update for $\bm{\delta}^{l-1}$ (\ref{delta_BP}).

\begin{eqnarray}
	\bm{\delta}^{l-1}&=&g_{l-1}^\prime(\mathbf{z}^{l-1})\odot[(\mathbf{W}^l)^T+2(\mathbf{A}^{l-1}\mathbf{Q}^l)^T]\bm{\delta}^l \nonumber \\
	\label{delta_BP}
\end{eqnarray}

(\ref{delta_BP}) for backpropagating $\bm{\delta}^l$ clearly reveals the symmetry between forward and backpropagation. The matrix $\mathbf{A}^{l-1}\mathbf{Q}^l$ which is the  most computationally costly part in forward propagation (\ref{FP}) appears again in transposed form in backpropagation (\ref{delta_BP}). Thus the large matrix $\mathbf{A}^{l-1}\mathbf{Q}^l$ can be cached during forward propagation and reused during backpropagation making this BP algorithm for QNNs computationally efficient.

Letting $\mathbf{V}^l=\mathbf{A}^{l-1}\mathbf{Q}^l$, the expression for back propagation can be rewritten as in (\ref{ueq}).

\begin{equation} \label{ueq}
	\bm{\delta}^{l-1}=g_{l-1}^\prime (\mathbf{z}^{l-1}) \odot [(\mathbf{W}^l)^T+2(\mathbf{V}^l)^T]\bm{\delta}^l
\end{equation}

To start backpropagation, $\bm{\delta}^{l}$ for the last layer namely $\bm{\delta}^{L}$ must first be computed.

For the last layer: 

\begin{eqnarray*}
	a^L_i &=& \hat{y}_i   \\
	\hat{y}_i &=& \sigma(z^L_i),  \because a_i^L=\sigma(z^L_i) \\
	\therefore \frac{\partial\hat{y}_I}{\partial z^L_i} &=& \hat{y}_i(1-\hat{y}_i)
\end{eqnarray*}

For the last layer, the Chain Rule yields,

\begin{equation}
	\delta^L_i = \frac{\partial\mathcal{L}}{\partial z^L_i} 
\end{equation}
\begin{eqnarray}
	&=& \frac{\partial\mathcal{L}}{\partial a^L_i}.\frac{\partial a^L_i}{\partial z^L_i}
	= \frac{\partial\mathcal{L}}{\partial\hat{y}_i}. \frac{\partial\hat{y}_i}{\partial z^L_i} \nonumber \\ 
	&=& \hat{y}_i-y_i. \label{vecto}
\end{eqnarray}

(\ref{vecto}) results because $a^L_i =\hat{y}_i =  -\left[\frac{y_i}{\hat{y}_i}-\frac{(1-y_i)}{(1-\hat{y}_i)}\right]\hat{y}_i(1-\hat{y}_i)$. (\ref{vecto}) can be written elegantly in vectorized form given in (\ref{dL}).

\begin{equation}\label{dL}
	\bm{\delta}^L = \mathbf{\hat{y}}-\mathbf{y}
\end{equation}

In \textbf{Algorithm} \ref{alg1} and \textbf{Algorithm} \ref{alg2}: $\mathbb{D} = \Set{ (\mathbf{x}^i,\mathbf{y}^i)| i = 1 \dots N  }$ is the training dataset.  Based on the above equations, the training algorithm for QNNs is presented in Algorithm \ref{alg1}.

\begin{algorithm}[H]
	\caption{QNN Training Algorithm}
	\begin{algorithmic}				
		\STATE {\textsc{TRAIN\_QNN}}$(\mathbb{D})$
		\STATE Initialize: $ \eta, \Set{\mathbf{Q}^l,\mathbf{W}^l,\mathbf{b}^l | l=1,2,\dots,L} $
		\STATE {\textbf{ Iterate till convergence}}
		\STATE \hspace{0.3cm} \% Forward Propagation
		\STATE \hspace{0.5cm} $ \mathbf{a}^1\leftarrow\mathbf{x}^i$
		\STATE \hspace{0.5cm} \textbf{ for } $ l = 1,2,...,L$
		\STATE \hspace{1cm} $\mathbf{V}^l\leftarrow\mathbf{A}^{l-1}\mathbf{Q}^l$ \% Cache for BP
		\STATE \hspace{1cm} $\mathbf{z}^l\leftarrow\mathbf{b}^l+(\mathbf{W}^l+\mathbf{V}^{l})\mathbf{a}^{l-1}$ \hspace{0.2cm}\% Cache for BP
		\STATE \hspace{1cm} $\mathbf{a}^l\leftarrow g(\mathbf{z}^l) $ \% Cache for BP
		\STATE \hspace{0.5cm} $ \mathbf{\hat{y}} \leftarrow \mathbf{a}^L$
		\STATE \hspace{0.5cm} $ \bm{\delta}^L \leftarrow \mathbf{\hat{y}}-\mathbf{y}^i$
		\STATE \hspace{0.3cm} \% Backward Propagation
		\STATE \hspace{0.5cm}\textbf{ for } $ l = L,L-1,\dots,2$
		\STATE \hspace{1cm} $ \bm{\delta}^{l-1}=g_{l-1}^\prime (\mathbf{z}^{l-1}) \odot [(\mathbf{W}^l)^T+2(\mathbf{V}^l)^T]\bm{\delta}^l $
		\STATE \hspace{0.5cm}\textbf{ for } $ l = 1,2,\dots,L$
		\STATE \hspace{1cm} $ \mathbf{b}^l\leftarrow\mathbf{b}^l-\eta\bm{\delta}^l$
		\STATE \hspace{1cm} $  \mathbf{W}^l\leftarrow\mathbf{W}^l-\eta\bm{\delta}^l\otimes\mathbf{a}^{l-1}$
		\STATE \hspace{1cm} $ \mathbf{Q}^l\leftarrow\mathbf{Q}^l-\eta\bm{\delta}^l\circledast\mathbf{M}(\mathbf{a}^{l-1})$
		\STATE \hspace{0.3cm} $i \leftarrow i+1$ 
		\STATE \textbf{ return}  $\Set{\mathbf{Q}^l,\mathbf{W}^l,\mathbf{b}^l | l=1,2,\dots,L} $
	\end{algorithmic}
	\label{alg1}
\end{algorithm}

Next a reduced parameter QNN model that provides a compromise between model complexity and representation power is proposed.

\section{Reduced Parameter Quadratic Neural Networks (RPQNN)}

One possible approach for reducing the number of parameters in the standard quadratic neuron model is to consider only quadratic functions that are product of the affine functions. In this model each neuron with $n$ inputs has only $2n$ additional parameters instead of $\frac{n(n+1)}{2}$. In the following a new Reduced Parameter Quadratic Neural Networks (RPQNN) model is proposed. \\

The output of layer $l$ in the RPQNN is calculated as follows:		

\begin{eqnarray}
	z_i^l &=& (W_{i:}\mathbf{a}^{l-1}+b_i^l)(U^l_{i:}\mathbf{a}^{l-1}+c_i^l) \\
	\mathbf{z}^l &=& (\mathbf{W}^l\mathbf{a}^{l-1}+\mathbf{b}^l)\odot(\mathbf{U}^l\mathbf{a}^{l-1}+\mathbf{c}^l)\\
	\mathbf{a}^l &=& g_l(\mathbf{z}^l) 
\end{eqnarray}

\begin{algorithm}[!h]
	\caption{RPQNN Training Algorithm}
	\begin{algorithmic}				
		\STATE {\textsc{TRAIN\_RPQNN}}($\mathbb{D}$)
		\STATE Initialize:  $\eta, \Set{\mathbf{W}^l,\mathbf{U}^l,\mathbf{b}^l,\mathbf{c}^l | l=1,2,\dots,L } $
		\STATE {\textbf{ Iterate till convergence}}
		\STATE \hspace{0.3cm} \% Forward Propagation
		\STATE \hspace{0.3cm} $ \mathbf{a}^1\leftarrow\mathbf{x}^i$
		\STATE \hspace{0.5cm} \textbf{for } $ l = 1,2,...,L$
		\STATE \hspace{1cm} $ \mathbf{v}_1^l\leftarrow(\mathbf{W}^l\mathbf{a}^{l-1}+\mathbf{b}^l)$ \hspace{0.2cm}\% Cache for BP
		\STATE \hspace{1cm} $ \mathbf{v}_2^l\leftarrow(\mathbf{U}^l\mathbf{a}^{l-1}+\mathbf{c}^l)$ \hspace{0.2cm}\% Cache for BP
		\STATE \hspace{1cm} $ \mathbf{z}^l\leftarrow \mathbf{v}^l_1\odot\mathbf{v}^{l}_2$	\hspace{0.2cm}\% Cache for BP
		\STATE \hspace{1cm} $ \mathbf{a}^l\leftarrow g(\mathbf{z}^l) $ \hspace{0.2cm}\% Cache for BP
		\STATE \hspace{0.5cm} $\mathbf{\hat{y}} \leftarrow \mathbf{a}^L$
		\STATE \hspace{0.5cm} $\bm{\delta}^L \leftarrow \mathbf{\hat{y}}-\mathbf{y}^i$
		\STATE \hspace{0.3cm} \% Backward Propagation
		\STATE \hspace{0.5cm}\textbf{ for } $ l = L,L-1,\dots,2$
		\STATE \hspace{1cm} $\bm{\delta}^{l-1}\leftarrow g^\prime_{l-1}(\mathbf{z}^{l-1}) \odot [(\mathbf{U}^l)^T(\bm{\delta}^l\odot \mathbf{v}_1^l) + (\mathbf{W}^l)^T(\bm{\delta}^l\odot \mathbf{v}_2^l)]$
		\STATE \hspace{0.5cm}\textbf{ for } $ l = 1,2,\dots,L$
		\STATE \hspace{1cm} $ \mathbf{b}^l\leftarrow\mathbf{b}^l-\eta\bm{\delta}^l\odot \mathbf{v}^l_1$
		\STATE \hspace{1cm} $ \mathbf{c}^l\leftarrow\mathbf{c}^l-\eta\bm{\delta}^l\odot \mathbf{v}^l_2$
		\STATE \hspace{1cm} $  \mathbf{W}^l\leftarrow\mathbf{W}^l-\eta(\bm{\delta}^l\odot \mathbf{v}_2^l)\otimes\mathbf{a}^{l-1}$
		\STATE \hspace{1cm} $  \mathbf{U}^l\leftarrow\mathbf{U}^l-\eta(\bm{\delta}^l\odot \mathbf{v}_1^l)\otimes\mathbf{a}^{l-1}$
		\STATE \hspace{0.3cm} $i \leftarrow i+1$ 
		\STATE \textbf{ return }  $\Set{\mathbf{W}^l,\mathbf{U}^l,\mathbf{b}^l,\mathbf{c}^l | l=1,2,\dots,L } $
	\end{algorithmic}
	\label{alg2}
\end{algorithm}

The parameters $\Theta$ are $w_{ij}^l$, $b^l_i$, $u^l_{ij}$ and $c^l_i$. \\

Using the Chain Rule (\ref{chain3}) : \\

\begin{equation} \label{chain3}
	\frac{\partial\mathcal{L}}{\partial w^l_{ij}}=\frac{\partial \mathcal{L}}{\partial z^l_i}\frac{\partial z^l_i}{\partial w_{ij}^l}=\delta^l_i\frac{\partial z^l_i}{\partial w^l_{ij}}
\end{equation}

\begin{equation} \nonumber
	\frac{\partial z^l_i}{\partial w^l_{ij}}=(U^l_{i:}\mathbf{a}^{l-1}+c_i^l)a_j^{l-1}
\end{equation}

\begin{equation} \label{chain4}
	\frac{\partial\mathcal{L}}{\partial w^l_{ij}}=\delta^l_i(U_{i:}^l\mathbf{a}^{l-1}+c_i^l)a_j^{l-1}
\end{equation}

Using the concept of matrix derivatives, the above weight update equations (\ref{chain4}) can be vectorized and presented concisely as follows (\ref{con}):

\begin{equation} \label{con}
	\frac{\partial\mathcal{L}}{\partial\mathbf{W}^l}=[\bm{\delta}^l\odot(\mathbf{c}^l+\mathbf{U}^l\mathbf{a}^{l-1})]\otimes \mathbf{a}^{l-1}
\end{equation}

By symmetry, we obtain the gradient with respect to $\mathbf{U}$ (\ref{usym}):

\begin{equation} \label{usym}
	\frac{\partial\mathcal{L}}{\partial\mathbf{U}^l}=[\bm{\delta}^l\odot(\mathbf{b}^l+\mathbf{W}^l\mathbf{a}^{l-1})]\otimes \mathbf{a}^{l-1}
\end{equation}

Next is the gradient of $\mathbf{b}$ which can be calculated as shown below (\ref{bup})

\begin{equation}  \label{bup}
	\frac{\partial \mathcal{L}}{\partial b^l_i}=\frac{\partial\mathcal{L}}{\partial z^l_i}\frac{\partial z^l_i}{\partial b^l_i}=\delta^l_i(c^l_i+U_{i:}\mathbf{a}^{l-1})
\end{equation}

(\ref{bup}) can be vectorized to yield (\ref{cup})

\begin{equation} \label{cup}
	\frac{\partial\mathcal{L}}{\partial\mathbf{b}^l}=\bm{\delta}^l\odot (\mathbf{c}^l+\mathbf{U}^l\mathbf{a}^{l-1})
\end{equation}

By symmetry with (\ref{cup}), the gradient of $\mathbf{c}$ is (\ref{cup1})

\begin{equation} \label{cup1}
	\frac{\partial\mathcal{L}}{\partial\mathbf{c}^l}=\bm{\delta}^l \odot (\mathbf{b}^l+\mathbf{W}^l\mathbf{a}^{l-1})
\end{equation}

Next we consider the equation for backpropagating $\bm{\delta}^l $. Using the multivariable Chain Rule:

\begin{eqnarray}
	\delta^{l-1}_i&=&\frac{\partial L}{\partial z_i^{l-1}} \nonumber\\
	&=&\sum_{k}^{n_l}\frac{\partial L}{\partial z_k^l} \frac{\partial z_k^l}{\partial a^{l-1}_i} \frac{\partial a^{l-1}_i}{\partial z^{l-1}_i} \label{eq5} 
\end{eqnarray}

(\ref{eq5}) can be simplified to obtain (\ref{d1}).

\begin{equation}\label{d1}
	\delta^{l-1}_i	= g_{l-1}^\prime (z_i^{l-1})\sum_{k=1}^{n_l}\delta_k^l\frac{\partial z^l_k}{\partial a^{l-1}_i}
\end{equation}

Now, $ \frac{\partial z^l_k}{\partial a^{l-1}_i} $ in (\ref{d1}) can be calculated as follows:

\begin{eqnarray}
	\frac{\partial z_k^l}{\partial a_i^{l-1}}&=&\frac{\partial}{\partial a_i^{l-1}} \left[(b_k^l+W_{k:}^l\mathbf{a}^{l-1})(c_k^l+U_{k:}^l\mathbf{a}^{l-1}) \right] \nonumber \\
	&=&\frac{\partial}{\partial a_i^{l-1}}\left[b_k^l(U_{k:}^l\mathbf{a}^{l-1})+c_k^l(W_{k:}\mathbf{a}^{l-1})\right. \nonumber \\
	&+&\left. (W_{k:}^l\mathbf{a}^{l-1})(U^l_{k:}\mathbf{a}^{l-1}) \right] \nonumber \\
	\frac{\partial z_k^l}{\partial a_i^{l-1}}&=& b^l_ku^l_{ki}+c_k^lw_{ki}^l+w_{ki}^l(U_{k:}^l\mathbf{a}^{l-1}) \nonumber\\
	&+&u_{ki}^l(W_{k:}^l\mathbf{a}^{l-1})
	\label{d2}
\end{eqnarray}

Substituting (\ref{d2}) in (\ref{d1}) we get:

\begin{eqnarray}
	\delta_i^{l-1}&=&g_{l-1}^\prime(z_i^{l-1})\sum_{k=1}^{n_l}\left[\delta_k^lb_k^lu_{ki}^l+\delta_k^lc_k^lw_{ki}^l\right.\nonumber\\
	&+& \delta_k^lw_{ki}^l(U_{k:}^l\mathbf{a}^{l-1})+ \delta_k^lu_{ki}^l(W_{k:}^l\mathbf{a}^{l-1}) ] 
	\label{d3}
\end{eqnarray}

(\ref{d3}) can be vectorized and expressed in compact and computationally efficient form (\ref{d4}).

\begin{eqnarray}
	\bm{\delta}^{l-1}&=&g_{l-1}^\prime(\mathbf{z}^{l-1}) \nonumber\\ &\odot& \left[ (\mathbf{U}^l)^T(\bm{\delta}^l\odot \mathbf{b}^l) \right.\nonumber \\ &+&(\mathbf{W}^l)^T(\bm{\delta}^l\odot\mathbf{c}^l)\nonumber \\ &+& (\mathbf{W}^l)^T(\bm{\delta}^l\odot \mathbf{U}^l\mathbf{a}^{l-1}) \nonumber \\ &+& \left. (\mathbf{U}^l)^T (\bm{\delta}^l \odot \mathbf{W}^l\mathbf{a}^{l-1})\right] \nonumber \\
	&=&g^\prime(\mathbf{z}^{l-1})\odot \left[ (\mathbf{U}^l)^T\left[\bm{\delta}^l\odot\mathbf{b}^l+\bm{\delta}^l
	\odot \mathbf{W}^l\mathbf{a}^{l-1} \right]\right. \nonumber \\
	&&\hspace{0.4cm}+ (\mathbf{W}^l)^T\left.\left[\bm{\delta}^l\odot\mathbf{c}^l+\bm{\delta}^l
	\odot \mathbf{U}^l\mathbf{a}^{l-1} \right]\right]  \nonumber \\
	&=& g^\prime (\mathbf{z}^{l-1}) \odot \left[ (\mathbf{U}^l)^T\left[\bm{\delta}^l\odot (\mathbf{W}^l\mathbf{a}^{l-1}+\mathbf{b}^l)\right]\right. \nonumber \\
	&&\hspace{0.4cm}+(\mathbf{W}^l)^T\left.\left[\bm{\delta}^l\odot ( \mathbf{U}^l\mathbf{a}^{l-1}+\mathbf{c}^l)\right]\right. \nonumber \\
	\bm{\delta}^{l-1}&=& g^\prime_{l-1}(\mathbf{z}^{l-1}) \odot \left\{ (\mathbf{U}^l)^T\left[\bm{\delta}^l\odot (\mathbf{W}^l\mathbf{a}^{l-1}+\mathbf{b}^l)\right]\right. \nonumber \\
	&&\hspace{1.4cm}+(\mathbf{W}^l)^T\left.\left[\bm{\delta}^l\odot ( \mathbf{U}^l\mathbf{a}^{l-1}+\mathbf{c}^l)\right]\right\}
	\label{d4}
\end{eqnarray}

The quantities $ \mathbf{W}^l\mathbf{a}^{l-1}+\mathbf{b}^l $ and $ \mathbf{U}^l\mathbf{a}^{l-1}+\mathbf{c}^l $ can be cached during Forward Propagation and reused during Backpropagation making this model computationally efficient. Based on the above equations, the training algorithm for RPQNN is presented in Algorithm \ref{alg2}.

\section{Computational Complexity of QNN and RPQNN}

\subsection{Time complexity of standard ANN}

\subsubsection{Forward Propagation}

In a standard ANN layer, matrix multiplication is the most expensive operation during forward propagation. Each floating point multiplication is assumed to require a fixed time and consume a fixed amount of energy on a given hardware platform. So the number of floating point multiplications is a measure of execution time as well as energy consumption. Since,

\begin{equation}  
	\mathbf{z}^l = \mathbf{W}^l\mathbf{a}^{l-1}+\mathbf{b}^l
\end{equation}

Where $\mathbf{W}^l$ is a $n_l\times n_{l-1}$ matrix, $\mathbf{a}^{l-1}$ is a $n_{l-1}$ dimension vector and $\mathbf{b}^l$ is a $n_l$ dimension vector. Thus $n_ln_{l-1}$  multiplications are performed in each layer during Forward Propagation. Thus the total number of multiplications needed for Forward Propagation in a standard ANN is: 

\begin{equation}
	A_{FP} = \sum_{l=1}^{L} n_ln_{l-1}
\end{equation}

\subsubsection{Backpropagation}
The computation of $\bm{\delta}^{l-1}$ from $\bm{\delta}^l$ requires a matrix multiplication and a Hadamard product. So $n_ln_{l-1}+n_{l-1}$ multiplications are needed.

\subsubsection{Parameter Update}
From Algorithm (\ref{alg1}),  we see that the $\mathbf{b}^l$ update  requires $n_l$ multiplications. Also from Algorithm (\ref{alg1}) we see that $\mathbf{W}^l$ update requires $n_ln_{l-1}+n_{l-1}$ multiplications. Thus the time complexity of updates is $n_ln_{l-1}+n_l+n_{l-1}$ multiplications.

Thus the overall time complexity for updating the parameters once in a standard ANN  using the Backpropagation Algorithm is:

\begin{equation}
	A_{BP} =\sum_{l=2}^{L} \left[2n_ln_{l-1}+n_l+2n_{l-1}\right] 			 				
\end{equation}

\subsection{Time Complexity of QNN}

\subsubsection{Forward Propagation}

In the following we examine the aditional multiplications needed during Forward Propagation in QNNs. Each quadratic neuron in the $l^{th}$ layer requires the computation of $\mathbf{a}^T\mathbf{Q}\mathbf{a}$ to calculate its output. $\mathbf{a}^T\mathbf{Q}\mathbf{a}$ requires $\frac{n_{l-1}(n_{l-1}+1)}{2}+n_{l-1}$ multiplications since $Q$ is symmetric. This is because the terms $q_{ij}x_i x_j$ and $q_{ji}x_jx_i$ can be reduced to a single term $q_{ij}x_ix_j$ by storing the entire coefficient value in $q_{ij}$ and setting $q_{ji}$ to be zero. Thus the extra computational cost is $n_l\left(\frac{n_{l-1}(n_{l-1}+1)}{2}+n_{l-1}\right)$.

\subsubsection{Backpropagation}

Since Algorithm \ref{alg1} uses the matrix $\mathbf{V}^l$ cached during Forward Propagation, the extra computation comes from $2(\mathbf{V}^l)^T$. Thus $n_{l-1}n_l$ extra multiplications are needed to compute $\bm{\delta}^{l-1}$ from $\bm{\delta}^l$. 

\subsubsection{Parameter Update}
The number of multiplications needed to update $\mathbf{b}^l$ and $\mathbf{W}^l$ are the same for QNN and ANN. 

To update each $Q_{lk}$, $\frac{n_{l}n_{l-1}(n_{l-1}+1)}{2}+n_ln_{l-1}$  extra multiplications are needed since $Q_{lk} = \delta^l_k M^{lk}$. Thus, to update every $Q^{lk}$, $\frac{n_{l}n_{l-1}(n_{l-1}+1)}{2}+n_ln_{l-1}$ multiplications are needed.

\begin{table}[h]
	\centering
	\caption{Extra floating point multiplications per layer}
	\begin{tabular}{|c|c|c|c|}  \hline
		\textbf{Algorithm} & \begin{tabular}{c}
			\textbf{Forward} \\
			\textbf{Propagation}
		\end{tabular}  & \textbf{BP} & \begin{tabular}{c}
			\textbf{Parameter} \\ \textbf{Update}
		\end{tabular}  \\ \hline
		QNN  & \begin{tabular}{c}
			$\frac{n_ln_{l-1}(n_{l-1}+1)}{2}$ \\
			$+n_ln_{l-1}$
		\end{tabular}  & $n_ln_{l-1}$ & \begin{tabular}{c}
			$\frac{n_{l}n_{l-1}(n_{l-1}+1)}{2}$ \\ $+n_ln_{l-1}$
		\end{tabular} \\ \hline
		RPQNN  & $n_ln_{l-1}+n_l$ & \begin{tabular}{c}
			$n_ln_{l-1}$ \\ $+2n_l$
		\end{tabular}  & \begin{tabular}{c}
			$3n_ln_{l-1}-$ \\
			$n_{l-1}+7n_l$ 
		\end{tabular}\\ \hline
	\end{tabular}
	\label{timec}
\end{table}
\setlength{\arraycolsep}{5pt}	
\subsection{Time Complexity of RPQNN}

From Algorithm \ref{alg2}, we see that the excess multiplications needed during Forward Propagation is due to an extra matrix multiplication and Hadamard product. Thus the number of extra multipliations needed during Forward Propagation in RPQNN is:

\begin{equation}
	R_{FP} = \sum_{l=1}^{L} \left[2n_ln_{l-1}+n_l\right].
\end{equation}

From Algorithm \ref{alg2}, we see that Back Propagation in RPQNN requires 2 extra Hadamard products and a matrix multiplication. Thus the number of extra multipliations needed during back Propagation in RPQNN is:

\begin{equation}
	R_{BP} =\sum_{l=2}^{L} \left[4n_ln_{l-1}+6n_l+n_{l-1}\right].
\end{equation}

Table \ref{timec} summarizes the additional floating point multiplications required per layer for QNN and RPQNN.

\subsection{Space Complexity of QNN and RPQNN}

\begin{table}
	\centering
	\caption{Excess memory reqirement per layer}
	\begin{tabular}{|c|c|}  \hline
		\textbf{Algorithm} & \textbf{Additional Memory Required Per Layer} \\ \hline
		QNN  & $n_l\frac{n_{l-1}(n_{l-1}+1)}{2}+2n_ln_{l-1}$  \\ \hline
		RPQNN &  $n_ln_{l-1}+3n_l$\\ \hline
	\end{tabular}
	\label{memory_c}
\end{table}

The standard ANN model requires the storage of $\mathbf{W}^l$, $\mathbf{b}^l$, $\mathbf{z}^l$ and $\mathbf{a}^l$ for each layer. From Algorithm \ref{alg1}, we see observe that QNN requires the additional storage of the sparse $\mathbf{Q}^l$ and non-sparse $\mathbf{V}^l$ matrices. The storage of $\mathbf{A}^{l-1}$ is not required since it can be constructed solely from already cached $\mathbf{a}^{l-1}$.  From Algorithm \ref{alg2}, we see that RPQNN requires the additional storage of $\mathbf{U}^l$, $\mathbf{c}^l$, $\mathbf{v}^l_1$, and $\mathbf{v}^l_2$. Based on the above considerations, the extra floating point storage required by QNN and RPQNN is summarized in Table \ref{memory_c}.

\subsection{Overall Asymptotic Complexity}

Table \ref{timec} can be simplified if we assume that $n_l$ and $n_{l-1}$ are approximately equal and use the standard Big $\mathcal{O}$ notation. From Table I it is clear that the asymptotic time-complexity of a QNN layer is $\mathcal{O}(n^3)$, whereas the time-complexity of a standard ANN or RPQNN layer is $\mathcal{O}(n^2)$. From Table II, we see that the asymptotic space-complexity per layer is $\mathcal{O}(n^3)$ for QNN and $\mathcal{O}(n^2)$ for both standard ANN and RPQNN. This polynomial complexity is acceptable since widely used practical algorithms like Gaussian Elimination has $\mathcal{O}(n^3)$ complexity. 

\section{Results and Discussion}  \label{results}	    	

In this section, we compare the performance of  QNNs and standard ANNs on the following 3 benchmark classification datasets:

\begin{enumerate}
	\item Nonlinear Cluster dataset
	\item MNIST dataset
\end{enumerate}

\begin{table}[!h]
	\caption{Nonlinear Cluster dataset}
	\centering
	\begin{tabular}{|c|c|c|c|} \hline
		\textbf{Color} & Red & Black & Magenta  \\ \hline
		\textbf{Mean} & (-16,0) & (-8,0) & (0,0) \\  \hline
		\textbf{Covariance} & $\begin{bmatrix} 
			1 & -0.3\\
			-0.3 & 1 \end{bmatrix}$&
		$\begin{bmatrix}
			1&0\\0&1
		\end{bmatrix}$ & 
		$\begin{bmatrix}
			1 & 0.3 \\0.3&1
		\end{bmatrix}$ \\ \hline
		\textbf{Color} & Green & Cyan & Blue \\ \hline
		\textbf{Mean} & (0,10) & (-8,10) & (-16,10)\\ \hline
		\textbf{Covariance} & $\begin{bmatrix}
			1&-0.3\\-0.3&1
		\end{bmatrix}$&
		$\begin{bmatrix}
			1&0\\0&1
		\end{bmatrix}$&
		$\begin{bmatrix}
			1 & 0.3 \\0.3&1
		\end{bmatrix}$ \\	\hline									
	\end{tabular}
	\label{cluster_parameters}
\end{table}

\begin{figure}[!h]
	\centering 
	\includegraphics[scale=0.4]{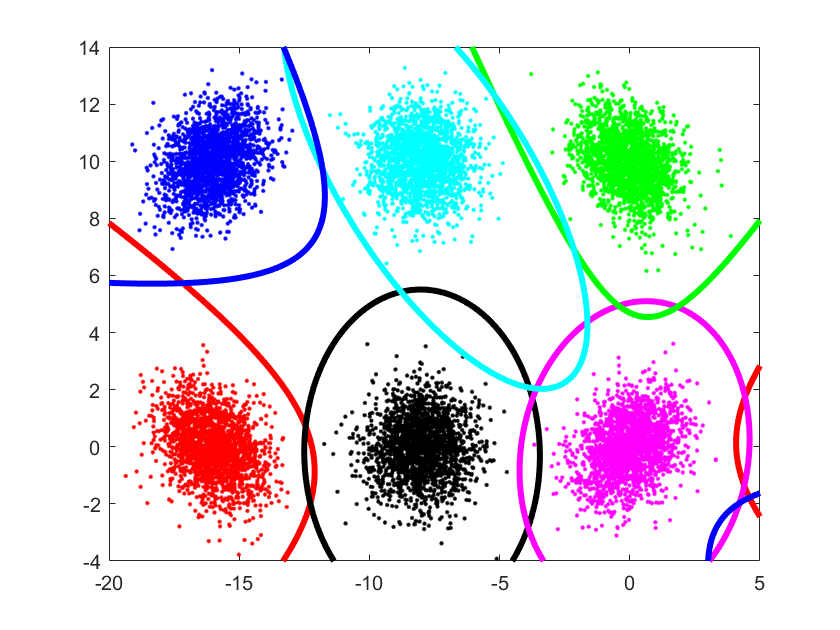}
	\caption{A dataset that is not linearly separable and consists of 6 clusters. A single-layer 6 neuron QNN is able to successfully learn this dataset. Different classes and the associated neuronal decision boundaries are shown in the same color.}
	\label{clusters}
\end{figure}		

\begin{table*}[]
	\centering
	\tiny
	\caption{Comparison of ANN, QNN and RPQNN models on MNIST}		
	\begin{tabular}{|c|c|c|c|c|c|c|c|c|} \hline 		
		\textbf{Data}&\textbf{Max}&\textbf{Hidden}& \multicolumn{2}{c|}{\textbf{ANN}} & \multicolumn{2}{c|}{\textbf{QNN}} &\multicolumn{2}{c|}{\textbf{RPQNN}}\\ \cline{4-9}
		\textbf{Size}&\textbf{Epochs}&\textbf{Neurons} &$\mu\pm\sigma$ & Best/Worst	& $\mu\pm\sigma$ & Best/Worst& $\mu\pm\sigma$ & Best/Worst  \\ \hline
		\multirow{15}{*}{600}	& \multirow{5}{*}{5}	&10&	10.08 $\pm$	0.35 & 11.35/9.58 &	72.68 $\pm$	3.16 &	77.38/65.51 & 23.51 $\pm$ 6.22 & 35.19/13.68 \\
		&	&15&	11.18 $\pm$	2.16 &	19.45/8.92 &	76.47 $\pm$	3.15 &	81.14/67.85 & 49.57 $\pm$ 5.85 & 60.47/33.80 \\
		&	&20&	12.80 $\pm$	5.72 &	35.98/8.92 &	77.79 $\pm$	2.98  &	83.73/71.40 & 55.33 $\pm$ 5.26 & 64.93/46.76 \\
		&	&25 &	16.69 $\pm$	6.11 &	32.44/9.74 &	78.56 $\pm$	2.65 &	82.54/69.61 & 58.86 $\pm$ 5.14 & 69.01/50.60 \\
		&	&30 &	22.49 $\pm$	7.26 &	34.60/9.64 &	79.31 $\pm$	3.41 &	84.02/69.29 & 60.52 $\pm$ 4.86 & 68.17/52.09 \\ \cline{2-9}
		&	\multirow{5}{*}{10}&10&	13.12 $\pm$ 5.04 &	28.69/8.92& 81.00 $\pm$ 2.59&	85.04/75.00 & 65.57 $\pm$ 5.46 & 75.96/55.54  \\
		&	&15& 38.81 $\pm$ 3.63 &	47.94/31.26 & 83.01 $\pm$ 2.49 & 85.83/73.84 &	73.82 $\pm$ 3.54 & 81.66/64.92 \\
		&	&20& 51.57 $\pm$ 5.13&	63.32/42.96 &83.61 $\pm$ 2.54 & 86.21/75.42&	76.03 $\pm$	3.41&	81.02/69.00 \\
		&	&25&58.83 $\pm$ 4.45&	67.83/49.45 & 83.29 $\pm$ 1.99 & 85.67/77.84 & 77.00 $\pm$ 2.42 &	80.57/71.16  \\	
		&	&30& 61.40 $\pm$ 3.43&	68.96/54.51& 83.63 $\pm$ 1.66 & 86.63/81.00 & 77.34 $\pm$ 2.89 & 82.67/72.90 \\		\cline{2-9}					
		& \multirow{5}{*}{20}	&10&62.30 $\pm$ 3.69&	69.48/56.20 & 83.66 $\pm$ 1.50 & 86.93/80.77 & 80.24 $\pm$	2.61 &	83.18/73.53 \\
		&	&15& 75.26 $\pm$ 2.95 & 80.81/67.91 & 84.91 $\pm$ 1.04	& 86.71/82.31&	82.88 $\pm$	1.60	& 85.62/79.92 \\
		&	&20	&79.18 $\pm$	1.90 & 81.60/73.61 & 85.22 $\pm$ 1.00 & 87.14/83.26 & 83.44 $\pm$ 1.64 &	85.19/79.61 \\
		&	&25& 80.96 $\pm$	1.62 &83.91/77.32&	85.99 $\pm$ 0.70 &	87.29/84.64 & 84.05 $\pm$ 1.76 &	86.89/80.29 \\
		&	&30& 81.53 $\pm$ 1.58 &	84.00/77.68 & 85.45 $\pm$ 0.91 &	87.03/83.45 & 83.60 $\pm$ 1.92 &	86.02/76.95\\ 	\hline													
		\multirow{15}{*}{1200}	& \multirow{5}{*}{5}	&10&	11.48 $\pm$	2.74 & 18.20/8.92 &	82.62 $\pm$	2.64 &	85.97/73.15 & 68.41 $\pm$ 5.29 & 78.78/57.20 \\
		&	&15&	39.92 $\pm$	3.19 &	44.91/32.65 &	85.18 $\pm$	1.75 &	87.75/79.62 & 74.07 $\pm$ 3.58 & 79.59/63.01 \\
		&	&20&	51.22 $\pm$	3.58 &	60.09/45.31 &	85.14 $\pm$	2.27  &	88.64/79.54 & 76.29 $\pm$ 3.95 & 82.83/64.50 \\
		&	&25 &	59.17 $\pm$	4.22 &	69.67/51.35 &	85.29 $\pm$	2.40 &	88.40/77.65 & 78.32 $\pm$ 3.46 & 83.05/71.71 \\
		&	&30 &	62.24 $\pm$	3.97 &	72.23/54.63 &	85.97 $\pm$	1.59 &	88.53/81.08 & 79.92 $\pm$ 2.14 & 83.34/75.03 \\ \cline{2-9}
		&\multirow{5}{*}{10}	&10&	60.11 $\pm$ 3.30 &	65.89/52.86 & 84.69 $\pm$ 1.50 &	86.87/81.46 & 81.29 $\pm$ 2.19 & 84.89/76.69  \\
		&	&15& 75.67 $\pm$ 2.81 &	80.74/68.47 & 86.51 $\pm$ 1.22 & 89.03/83.84 & 83.70 $\pm$ 2.68	& 86.64/75.47 \\
		&&20& 79.69 $\pm$ 1.84 & 83.33/75.59 & 86.97 $\pm$ 1.46 & 88.98/83.18 & 84.46 $\pm$ 2.50	&	87.81/76.95 \\
		&& 25& 81.52 $\pm$ 1.93 & 84.36/76.06 & 87.27 $\pm$ 1.28 & 89.61/84.58 & 84.85 $\pm$ 1.71 &	87.39/81.58  \\	
		&&30& 82.50 $\pm$ 1.51 &	85.24/79.16 & 87.58 $\pm$ 1.25 & 89.44/83.56 & 85.48 $\pm$ 2.03 & 87.77/78.97 \\		\cline{2-9}					
		&\multirow{5}{*}{20}	&10&83.20 $\pm$ 1.27 &	85.55/80.13 & 85.66 $\pm$ 1.62 & 87.78/80.68 & 84.67 $\pm$	1.41 &	86.33/81.99 \\
		&	& 15 &86.30 $\pm$ 1.03 & 87.73/83.74 & 87.54 $\pm$ 0.88	& 88.65/85.21 &	85.16 $\pm$	2.04	& 87.35/79.79 \\
		&	&20	&87.27 $\pm$ 0.80 &88.34/85.20& 88.20 $\pm$ 0.70 & 89.62/86.57 &86.55 $\pm$ 2.33&	89.27/77.32 \\
		&	&25&87.64 $\pm$	0.42 & 88.41/86.62 & 88.45 $\pm$ 0.62 &	89.46/87.33 & 87.12 $\pm$ 1.22 &	88.91/84.61 \\
		&	&30& 87.97 $\pm$ 0.65 &	88.83/86.12 & 88.83 $\pm$ 0.72 &	89.85/86.50 & 86.53 $\pm$ 2.37 &	89.04/77.19\\ 	\hline			
		\multirow{15}{*}{6000}	& \multirow{5}{*}{5}	&10&	86.33 $\pm$	1.69 & 88.60/80.28 &	88.39 $\pm$	1.20 &	90.23/85.66 & 87.19 $\pm$ 1.15& 89.07/84.40 \\
		&	&15&	88.75 $\pm$	0.54 &	89.92/87.57 &	90.53 $\pm$	0.95 &	91.74/88.21 & 89.25 $\pm$ 1.22 & 90.82/86.03 \\
		&	&20& 89.39 $\pm$	0.52 &	90.19/88.20 &	91.49 $\pm$	0.81  &	92.47/88.58 & 89.87 $\pm$ 1.26 & 91.35/85.43 \\
		&	&25 &	89.70 $\pm$	0.35 &	90.27/88.84 &	91.81 $\pm$	0.78 &	92.79/89.50 & 90.41 $\pm$ 0.90 & 91.88/88.44 \\
		&	&30 &	90.03 $\pm$	0.38 &	90.61/89.21 &	91.91 $\pm$	0.70 &	92.94/90.03 & 90.60 $\pm$ 1.09 & 91.69/86.37 \\ \cline{2-9}
		& \multirow{5}{*}{10}	&10&	88.28 $\pm$ 1.03 &	90.02/86.10 & 89.40 $\pm$ 1.05 &	91.00/86.46 & 88.68 $\pm$ 0.65	& 89.96/87.58  \\
		&	&15&90.15 $\pm$ 0.48&	90.82/89.07 & 90.88 $\pm$ 0.89 & 91.89/88.26 &	90.41 $\pm$ 0.68 & 91.36/88.61 \\
		&	&20& 90.82 $\pm$ 0.37 &	91.55/90.06 & 91.98 $\pm$ 0.88 & 95.76/93.92& 90.99 $\pm$ 0.86 &	92.15/89.11 \\
		&	&25& 90.97 $\pm$ 0.45 &	91.71/90.00 & 92.69 $\pm$ 0.76 & 93.69/90.70 & 91.55 $\pm$ 0.97 & 92.79/89.14  \\	
		&	&30& 91.24 $\pm$ 0.47 &	91.96/90.30 & 93.04 $\pm$ 0.39 & 93.65/92.15 & 91.79 $\pm$ 1.00 & 92.97/87.97 \\		\cline{2-9}					
		& \multirow{5}{*}{20}	&10& 88.87 $\pm$ 0.66 &	90.14/87.61 & 89.32 $\pm$ 1.10 & 90.74/85.44 & 88.68 $\pm$	0.94 &	90.14/86.43 \\
		&	&15& 90.61 $\pm$ 0.45 & 91.40/89.44 & 91.74 $\pm$ 0.35	& 92.28/90.84 &	90.27 $\pm$	0.75 & 91.48/88.60 \\
		&	&20	& 91.35 $\pm$ 0.37 & 91.95/90.57 & 92.78 $\pm$ 0.41 & 93.73/92.11 & 91.31 $\pm$  0.86 &	92.34/88.74 \\
		&	&25&91.84 $\pm$	0.42 & 92.38/90.44 &	93.36 $\pm$ 0.39 &	93.96/92.35 & 91.50 $\pm$ 0.93	&	92.70/89.22 \\
		&	&30& 92.07 $\pm$ 0.48 &	92.80/90.88 &93.56 $\pm$ 0.45 &	94.41/92.22 & 92.08 $\pm$ 0.79 & 93.14/89.90\\  \hline													
		\multirow{15}{*}{60000}	&\multirow{5}{*}{5}	&10&	90.87 $\pm$	0.49 & 92.06/89.88 &	92.28 $\pm$	0.42 &	93.09/91.57 & 91.36 $\pm$ 0.36 & 91.98/90.69 \\
		&	&15&	92.66 $\pm$	0.38 &	93.67/91.85 &	94.21 $\pm$	0.49 &	95.01/92.84 & 93.14 $\pm$ 0.56 & 94.04/90.97 \\
		&	&20& 93.78 $\pm$ 0.42 &	94.54/92.82 &	95.30 $\pm$	0.40  &	96.02/94.16 & 94.19 $\pm$ 0.39 & 95.06/93.40 \\
		&	&25 & 94.44 $\pm$	0.24 &	94.73/93.81 &	95.96 $\pm$	0.18 &	96.24/95.59 & 94.62 $\pm$ 0.68 & 95.38/91.82 \\
		&	&30 &	94.76 $\pm$	0.27 &	95.22/94.12 &	96.35 $\pm$	0.31 &	96.85/95.81 & 95.22 $\pm$ 0.51 & 95.77/93.18 \\ \cline{2-9}
		& \multirow{5}{*}{10}	&10& 91.30	$\pm$ 0.46 &	92.16/90.37 & 92.41 $\pm$ 0.62 &	93.23/90.52 & 91.77 $\pm$ 0.50	& 92.41/90.46  \\
		&	&15& 93.14 $\pm$ 0.39 &	93.80/92.49 & 94.38 $\pm$ 0.43 & 94.96/93.07 &	93.53 $\pm$ 0.37	& 94.28/92.65 \\
		&	&20& 94.24 $\pm$ 0.27 &	94.81/93.82 & 95.54 $\pm$ 0.33 & 96.14/95.00 &	94.48 $\pm$	0.31 & 95.19/93.83 \\
		&	&25& 94.96 $\pm$ 0.29 &	95.62/94.42 & 96.07 $\pm$ 0.36 & 96.78/94.88 & 95.02 $\pm$ 0.52 &	95.70/93.78  \\	
		&	&30& 95.39 $\pm$ 0.21 &	95.86/94.90 & 96.49 $\pm$ 0.20 & 96.83/96.06 & 95.75 $\pm$ 0.28 & 96.39/95.28 \\		\cline{2-9}					
		& \multirow{5}{*}{20}	&10& 91.49 $\pm$ 0.47 &	92.14/90.33 & 92.70 $\pm$ 0.46 & 93.57/91.71 & 91.96 $\pm$ 0.57 &	92.83/90.13 \\
		&	&15& 93.37 $\pm$ 0.41 & 94.27/92.73 & 94.58 $\pm$ 0.29	& 95.12/93.63 &	93.59 $\pm$	0.46 & 94.51/92.10 \\
		&	&20	& 94.49 $\pm$ 0.28 & 94.97/93.93 & 95.57 $\pm$ 0.30 & 96.04/94.89 & 94.58 $\pm$ 0.40 & 95.16/93.43 \\
		&	&25& 95.19 $\pm$ 0.21 & 95.54/94.77 &	96.02 $\pm$ 0.49 &	96.47/93.98 & 95.23 $\pm$ 0.43 & 95.82/94.19 \\
		&	&30& 95.66 $\pm$ 0.22 &	96.06/95.33 & 96.54 $\pm$ 0.22 &	96.88/95.99 & 95.71 $\pm$ 0.31 & 96.19/95.05 \\ 
		\hline													
	\end{tabular}
	\label{mnist}
\end{table*}

The final classification test-accuracy is a random variable due to the random weight and bias initialization at the start of training. Due to the random parameter initialization, gradient descent starts at a different point in the parameter space and reaches a possibly different local minimum every time the model is trained. So, in the following, the average accuracy over 25 independent training sessions starting from initial random weights and biases is considered to average out the variation in test-accuracy due to the random initialization.   

\subsection{Performance on a Nonlinear Cluster dataset}\label{NCD}

Fig. \ref{clusters} shows a 6-cluster linearly non-separable dataset. The clusters were generated using 2D Gaussian random variables with the mean and covariance matrices given in Table \ref{cluster_parameters}. The Nonlinear Cluster dataset consists of 6 classes and a training dataset consisting of 2000 training pairs per class. The test dataset consists of 3000 training pairs with 500 pairs for each of the 6 classes. A single layer QNN consisting of 6 quadratic neurons was trained using (\ref{bupdate}), (\ref{w_update}) and (\ref{Q_update}) for 10000 epochs with a learning rate of 0.0001 Nonlinear Cluster dataset. With the above settings, the test-accuracy was observed to be $99.97\%$. Fig. \ref{clusters} shows the different classes and the decision boundaries of the 6 neurons in the single layer QNN in different colors. This dataset demonstrates that single layer QNNs can learn complex nonlinear quadric boundaries that a standard single layer ANN cannot learn. The decision boundaries of the QNN in this 2-dimensional example are ellipses and hyperbolas. In $n$-dimension the decision boundaries of the QNN will be general quadric surfaces of the form ($ \mathbf{x}^T\mathbf{Q}\mathbf{x}+\mathbf{w}^T\mathbf{x}+b = 0$). An hyper-ellipsoidal decision boundary can always be found such that $ \mathbf{x}^T\mathbf{Q}\mathbf{x}+\mathbf{w}^T\mathbf{x}+b > 0$ for inputs belonging to a bounded cluster and $\mathbf{x}^T\mathbf{Q}\mathbf{x}+\mathbf{w}^T\mathbf{x}+b < 0$ for inputs not belonging to the cluster. Since the boundary of a quadratic neuron can be an arbitrary hyper-ellipsoid it is clear that any bounded $C$ clusters dataset requires only a single layer QNN with $C$ neurons. This problem clearly demonstrates that single layer QNNs can solve problems that can only be solved using standard ANNs with hidden layers.

\subsection{Performance on the MNIST benchmark}

In the following, the performance of QNN, RPQNN and standard ANN models is compared on the widely used MNIST benchmark dataset \cite{MNIST}. MNIST provides a training set of 60,000 labelled 28$\times$28 pixel images of the 10 handwritten digits (0 to 9) and a test dataset of 10,000 images. A simple 2-layer feedforward ANN model was considered.  All models had 784 inputs (flattened 28 by 28 pixel images), a single hidden layer composed of logistic sigmoidal neurons and a 10 neuron output layer. For the ANN model, the output layer consisted of 10 logistic sigmoidal neurons, and for the QNN and RPQNN models, the output layer consisted of 10 sigmoidal neuron QNN and RPQNN layer, respectively. The different models were trained with the SGD algorithm, and a learning rate of 0.01 was used. Although MNIST contains 60,000 training examples, it is interesting to consider the performance of different models when trained with small subsets of MNIST. Table \ref{mnist} compares the mean accuracy achieved by different models under different training conditions. In particular Table \ref{mnist} is useful in identifying models that work well with small training sets and fewer hidden layer neurons.  These results indicate that when the number of hidden layer neurons is large the learned features become easily linearly separable and the complex quadric decision boundaries of quadratic neurons are not needed to separate the classes. Table \ref{mnist} clearly shows that QNN and RPQNN significantly outperform the standard ANN model when the number of hidden layer neurons, epochs and training dataset size are small.

\begin{table}[t]
\centering
\caption{Execution Time Vs Accuracy}
\scriptsize
\begin{tabular}{|c|c|c|c|c|}  \hline
	\textbf{Dataset} & \textbf{Metric} & \textbf{ANN} & \textbf{QNN} & \textbf{RPQNN} \\ \hline
	\multirow{2}{*}{MNIST} & Time  & 1.63 $\pm$ 0.04 s & 5.98 $\pm$ 0.05 s & 1.71 $\pm$ 0.01 s \\ \cline{2-5} 
	& Accuracy	& 94.76$\%$ & 96.35$\%$ & 95.22$\%$ \\ \hline 
\end{tabular}
\end{table}

\section{Conclusion}

This paper presented the derivation of elegant vectorized equations for QNNs and a new reduced parameter RPQNN model for the first time. Algorithm \ref{alg1} and Algorithm \ref{alg2} proposed in this paper allow the efficient implementation of QNNs and allow further theoretical study of QNN properties like gradient flow. The paper formulates QNN algorithms using matrix multiplication, outer product, Hadamard product and Kronecker products. All of the above operations have efficient implementations in standard linear algebra libraries. This paper also proved that the results of Forward Propagation can be cached and reused during Back Propagation in QNNs and RPQNNs resulting in efficient computational models.

The paper explored the advantages of using quadratic neurons in feedforward neural networks. Quadratic neurons are sparse in terms of parameters compared to other higher order neurons because every quadratic form can be represented by a symmetric matrix. Efficient vectorized update equations for a new Quadratic Regression model were presented and single quadratic neurons were shown to possess the ability to learn the XOR function like recently discovered human neocortical pyramidal neurons involved in higher order functions \citep{Gidon}. The BP algorithm for QNNs in matrix form clearly revealed the symmetry between forward and backpropagation (matrices occur as transposes). This paper showed that any dataset consisting only of bounded clusters can be classified efficiently by a single layer QNN. This paper proved that the number of quadratic neurons needed to classify a dataset consisting of $C$ bounded clusters is exactly $C$. 

A typical ANN model consists of multiple hidden layers that hierarchically extract a small compressed set of linearly separable features from a high dimensional input vector followed by a final Softmax layer that can perform only linear classification. Since a single QNN/RPQNN can have complex quadric decision boundaries, linear separability at the final layer is not needed, allowing the model to have fewer hidden layers. Thus the use of quadratic neurons only in the final few layers as needed can help achieve higher accuracies without introducing many extra parameters. The QNN and RPQNN models were shown to significantly outperfrom standard ANNs on the MNIST benchmark. In particular, QNN and RPQNN models are advantageous when the dataset size, training epochs and number of hidden layer neurons are small. Results indicate that a final layer of quadratic sigmoidal neurons can significantly reduce the number of hidden layer neurons in ANNs. 

A reduced parameter QNN called RPQNN architecture was proposed and shown to provide almost the same performance benefits as QNNs while being only slightly more costlier than ANNs. Theoretical and empirical comparisons of the computational complexity of standard ANNs and proposed QNN and RPQNN models were presented. 

The QNN model is shown to have an asymptotic computational complexity of $\mathcal{O}(n^3)$ per layer same as the famous Gaussian Elimination algorithm. The asymptotic complexity of both ANN and RPQNN are same $(\mathcal{O}(n^2))$, although experimental results indicate that RPQNN takes slightly more time in practice. Future work will explore the possible  advantages using quadratic neurons in recurrent neural networks.

\bibliographystyle{plain}

\end{document}